# BALTIC: A Benchmark and Cross-Domain Strategy for 3D Reconstruction Across Air and Underwater Domains Under Varying Illumination

**Michele Grimaldi[1], David Nakath[2], Oscar Pizarro[3], Jonatan Scharff Willners[4], Ignacio Carlucho[1], Yvan R. Petillot[1]**

**Abstract**

Robust 3D reconstruction across varying environmental conditions remains a critical challenge for robotic perception, particularly when transitioning between air and water. To address this, we introduce BALTIC, a controlled benchmark designed to systematically evaluate modern 3D reconstruction methods under variations in medium and lighting. The benchmark comprises 13 datasets spanning two media (air and water) and three lighting conditions (ambient, artificial, and mixed), with additional variations in motion type, scanning pattern, and initialization trajectory, resulting in a diverse set of sequences. Our experimental setup features a custom water tank equipped with a monocular camera and an HTC Vive tracker, enabling accurate ground-truth pose estimation. We further investigate cross-domain reconstruction by augmenting underwater image sequences with a small number of in-air views captured under similar lighting conditions. We evaluate Structure-from-Motion reconstruction using COLMAP in terms of both trajectory accuracy and scene geometry, and use these reconstructions as input to Neural Radiance Fields and 3D Gaussian Splatting methods. The resulting models are assessed against ground-truth trajectories and in-air references, while rendered outputs are compared using perceptual and photometric metrics. Additionally, we perform a color restoration analysis to evaluate radiometric consistency across domains. Our results show that under controlled, texture-consistent conditions, Gaussian Splatting with simple preprocessing (e.g., white balance correction) can achieve performance comparable to specialized underwater methods, although its robustness decreases in more complex and heterogeneous real-world environments.

**Keywords**

Marine Robotics, Structure-from-Motion, Underwater Vision, 3D Representations

## Introduction

High-quality 3D reconstruction underpins mapping, localization, navigation, and scene understanding across computer vision, robotics, archaeology, and geoscience. Recent advances in Structure-from-Motion (SfM), Neural Radiance Fields (NeRF), and 3D Gaussian Splatting (3DGS) have significantly improved reconstruction fidelity and rendering quality (Ren et al. 2024; Ress et al. 2025; Kopanas and Drettakis 2023). However, these methods are predominantly evaluated in controlled, in-air environments with stable illumination, typically under natural lighting conditions. This evaluation setting overlooks a critical class of applications in underwater and cross-medium (air–water) environments, which are central to marine robotics, underwater archaeology, and infrastructure inspection. In such scenarios, visual degradation is severe: light scattering and absorption reduce contrast and color consistency, refraction introduces geometric distortions, and illumination varies significantly between directional, artificial, and mixed sources. These factors jointly challenge both geometric reconstruction and radiometric consistency, often leading to failure modes that are not observed in standard in-air benchmarks (Grimaldi et al. 2023). Despite these challenges, existing datasets and benchmarks remain limited. Most isolate individual factors such as lighting or motion, and do not provide controlled variation across both medium and illumination, making it difficult to systematically evaluate reconstruction robustness or compare methods under realistic underwater conditions. As a result, the performance and limitations of modern reconstruction pipelines in such environments remain insufficiently understood. To address this gap, we introduce BALTIC, an open-source benchmark designed for controlled evaluation of 3D reconstruction across air and underwater domains. Our experimental setup consists of a custom water tank with a monocular camera enclosed in a dome port,

[1]School of Engineering & Physical Sciences, Heriot-Watt University, Edinburgh, UK
[2]Marine Data Science Group, Christian Albrechts University of Kiel, Kiel, Germany
[3]Department of Marine Technology, Norwegian University of Science and Technology (NTNU), Trondheim, Norway
[4]Frontier Robotics, The National Robotarium, Edinburgh, UK

**Corresponding author:**
Michele Grimaldi, Heriot-Watt University School of Engineering & Physical Sciences, Edinburgh, UK

coupled with a motion tracking system that provides accurate ground-truth trajectories. The benchmark comprises 13 datasets that systematically vary across two media (air and water), three lighting conditions (directional, artificial, and mixed), and multiple motion patterns, including structured scanning and free-form trajectories. This design enables repeatable yet diverse evaluation scenarios that emulate real-world amphibious and underwater deployments. Building on this benchmark, we evaluate a standard reconstruction pipeline based on COLMAP (Schönberger and Frahm 2016), and use its outputs as input to Neural Radiance Fields (NeRF) (Mildenhall et al. 2020) and 3D Gaussian Splatting (3DGS) (Kerbl et al. 2023). We assess performance in terms of trajectory accuracy, geometric completeness, and image fidelity using metrics such as Structured Similarity Index Metric(SSIM), Learned Perceptual Image Patch Similarity (LPIPS), and Peak Signal-to-Noise Ratio (PSNR), enabling a unified comparison across methods and conditions. This evaluation provides a critical analysis of the robustness and failure modes of current reconstruction approaches under underwater and mixed-domain conditions. In addition, we investigate cross-domain reconstruction, motivated by practical workflows in which objects are scanned in air prior to deployment underwater. We show that augmenting underwater sequences with a small number of in-air images captured under similar lighting conditions significantly improves reconstruction quality (Fig. 1), providing both geometric structure and radiometric priors. This result highlights a simple yet effective strategy for improving reconstruction reliability in degraded visual environments. Our contributions are threefold:

- **Open-source benchmark (BALTIC):** A controlled dataset for evaluating 3D reconstruction across air and underwater domains, with systematic variation in lighting, medium, and motion, and accurate ground-truth trajectories.
- **Comprehensive evaluation and comparison:** A systematic assessment of SfM, NeRF, and 3D Gaussian Splatting methods, including a critical analysis of their robustness and failure modes under underwater and mixed-lighting conditions.
- **Cross-domain reconstruction analysis:** A study showing that incorporating a small number of in-air images significantly improves underwater reconstruction, providing a practical strategy for real-world applications.

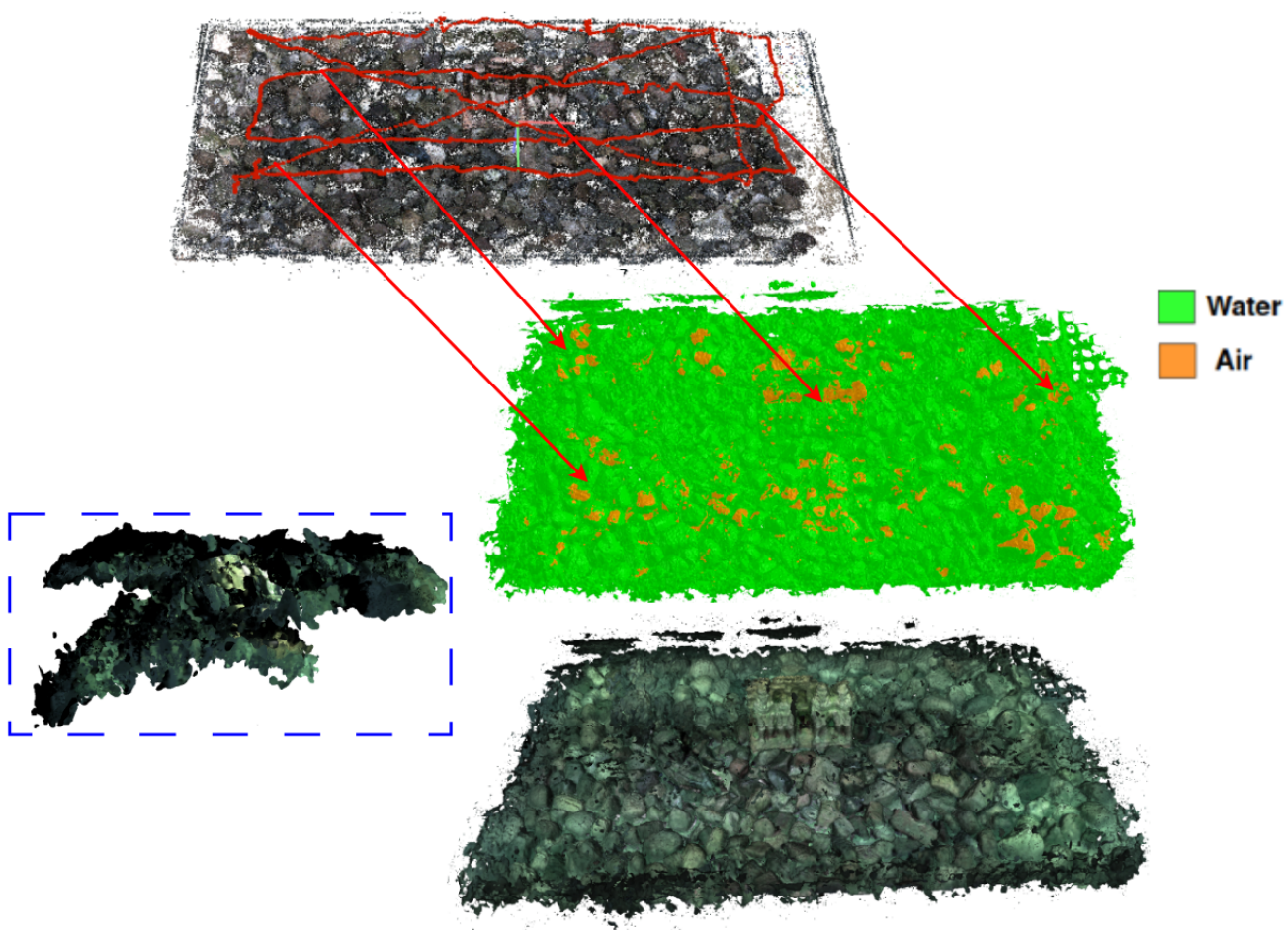

**Figure 1.** Cross-domain reconstruction. (Blue Rectangle) Failure with only E12 underwater images. Success using E12 underwater plus a few E6 in-air images. Color-coded mesh: underwater (green) vs. in-air (orange), from COLMAP Poisson reconstruction.

| Dataset | Description |
|---|---|
| E1 | Air-Dir-Lawn- |
| E2 | Air-Dir-Lawn+Init |
| E3 | Air-Dir-Free- |
| E4 | Air-Mix-Lawn+Init |
| E5 | Air-Mix-Free- |
| E6 | Air-Art-Lawn+Init |
| E7 | Air-Art-Free- |
| E8 | Wat-Dir-Lawn+Init |
| E9 | Wat-Dir-Free- |
| E10 | Wat-Mix-Lawn+Init |
| E11 | Wat-Mix-Free- |
| E12 | Wat-Art-Lawn+Init |
| E13 | Wat-Art-Free- |

**Table 1.** Overview of datasets E1–E13. Codes indicate: Medium: Air/Wat; Lighting: Dir (directional), Mix (mixed), Art (artificial); Trajectory: Lawn (lawn-mower), Free (free 3D); +Init: initialization maneuver.

## Related Work

3D reconstruction has been extensively studied across a range of environments, with methods broadly categorized into geometric approaches such as Structure-from-Motion (SfM), and learning-based methods such as Neural Radiance Fields (NeRF) and 3D Gaussian Splatting (3DGS). While these approaches have demonstrated strong performance in controlled, in-air settings, their behavior under challenging conditions such as underwater environments or varying illumination remains less well understood. SfM methods estimate camera poses and sparse geometry from feature correspondences, and often serve as the foundation for downstream reconstruction pipelines. In contrast, NeRF-based methods model scenes as continuous volumetric radiance fields, relying heavily on accurate camera poses and photometric consistency. More recently, 3D Gaussian Splatting has emerged as an efficient alternative, enabling real-time rendering with competitive reconstruction quality. Each of these approaches exhibits different sensitivities to visual degradation, including reduced contrast, scattering, and illumination changes. In the following, we review these three families of methods with a focus on their applicability and limitations in underwater and cross-domain scenarios.

### *Structure-from-Motion*

SfM reconstructs sparse 3D geometry by estimating camera motion and triangulating feature correspondences. Although effective in terrestrial settings, its robustness underwater remains underexplored, as color absorption, scattering, turbidity, motion blur, refraction and lighting variability complicate feature detection and matching. Previous studies (Lochhead and Hedley 2022) evaluated SfM underwater, but rarely under systematically varied illumination. Among existing pipelines, COLMAP (Schönberger and Frahm 2016) remains the de facto standard for accuracy and modularity, while alternatives such as GLOMAP (Pan et al. 2024) trade geometric precision for scalability and speed. In our work, we explicitly ensure that refraction is physically resolved by using a centered dome-port camera with underwater calibration. This setup allows us to focus solely on radiometric challenges rather than geometric distortions. For scenarios involving refracted data, refractive SfM approaches such as (Jordt-Sedlazeck and Koch 2013) or (She et al. 2022) provide suitable solutions. To enable consistent cross-domain evaluation, we undistort both in-air and underwater imagery from their respective camera models into a joint canonical pinhole representation. This common projection model ensures that the differences in reconstruction arise from medium- and illumination-related degradations rather than geometric incompatibilities.

### *Neural Radiance Fields*

Neural Radiance Fields (NeRF) (Mildenhall et al. 2020) learn continuous volumetric scene representations from SfM-posed multi-view images, enabling dense, photorealistic reconstruction. NeRF assumes consistent radiance and precise calibration, which are often violated underwater due to scattering, absorption, lighting variability, and noisy motion estimates. Extensions such as Mip-NeRF (Barron et al. 2021) improve multiscale generalization and reduce aliasing, while frameworks like Nerfstudio (Tancik et al. 2023) (with Nerfacto) balance speed, quality, and robustness. Underwater adaptations include WaterNeRF (Sethuraman et al. 2023) which integrates a physics-based image formation model into NeRF by explicitly estimating attenuation and backscatter to correct degraded views and recover depth, while SeaThru-NeRF (Levy et al. 2023) disentangles object and medium contributions, jointly learning scene radiance and medium properties to enable realistic underwater rendering and medium-free *in-air* reconstructions. Additional work addresses extreme lighting: RawNeRF (Mildenhall et al. 2021) leverages raw sensor data to preserve dynamic range, and relative illumination fields (She et al. 2025) capture spatially varying lighting in scattering media. Despite these advances, systematic evaluation of NeRF under controlled underwater and lighting variations remains limited.

### *3D Gaussian Splatting*

3D Gaussian Splatting (Kerbl et al. 2023) is a recent alternative to NeRF that uses explicit 3D Gaussians instead of volumetric representations, enabling faster training and real-time rendering. Like NeRF, it relies on accurate camera poses, typically from SfM pipelines, and may offer greater robustness to sparse views or pose drift. Its performance under challenging conditions, especially underwater scenes with scattering, absorption, and variable lighting, remains largely unexplored. Two recent works have applied 3DGS underwater: WaterSplat (Li et al. 2025) and SeaSplat (Yang et al. 2025). Both combine 3DGS with underwater image formation models to capture geometry and scattering in real time. WaterSplat separates geometry and medium into explicit and volumetric components, while SeaSplat constrains 3DGS using a physically grounded image formation model. Both show strong results on SeaThru-NeRF datasets and their own datasets; however, evaluations assume fixed exposure and static illumination. As with SeaThru-NeRF, WaterSplat, and SeaSplat, we do not perform a full radiometric evaluation on our datasets, such as using a MacBeth color checker (Pascale 2006). While a physically-based image formation model could potentially solve image restoration, neither these methods nor our work perform such evaluations. Consequently, we treat the applied adjustments as image enhancement aimed at supporting a downstream task, specifically geometric reconstruction.

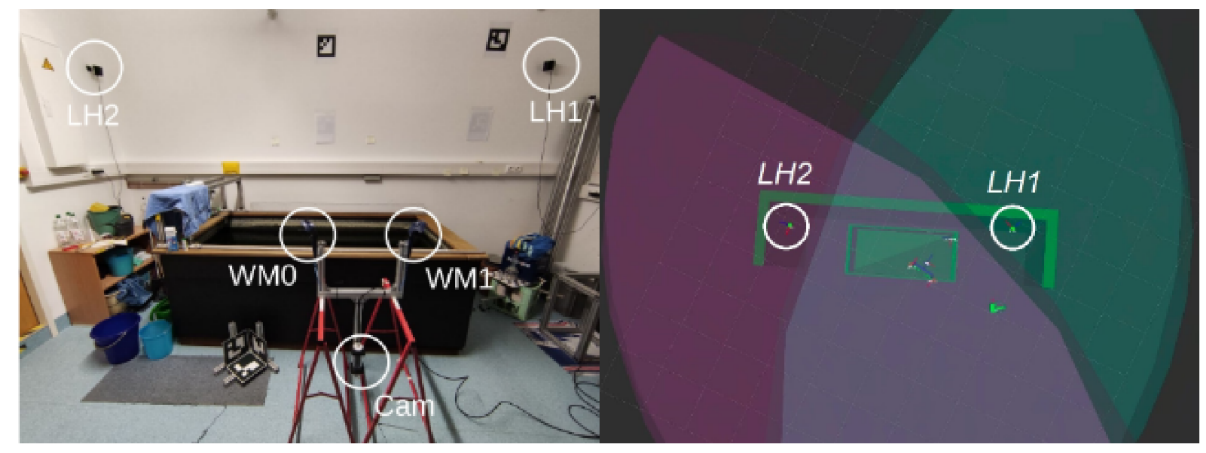

**Figure 2.** (Left) Side view of the real environment. (Right) Top view showing the HTC Vive system, with the lighthouses (LHs) covering the area and the WMs representing the Vive controllers equipped with IMUs (Winkel et al. 2023)

In our dataset construction, we emulate the characteristic appearance of the Baltic Sea by introducing a controlled mixture of dye to reproduce its greenish coloration and adding Maalox (Laux et al. 2002) as a scattering agent. This follows established experimental protocols for optical water tank studies and provides a simple and reproducible proxy for turbidity and light scattering, enabling systematic control of underwater visibility conditions for reconstruction experiments. While such handcrafted mixtures are effective for controlled benchmarking, more recent work on procedural underwater appearance modelling and optical ocean recipe generation provides more physically grounded and flexible alternatives to recreate complex water body characteristics (Schöntag et al. 2025).

## Dataset creation

We constructed a $2.2 \times 1 \times 0.8$ m water tank illuminated by three 50 W Wasler daylight bulbs (5400 K) fitted with Walimex diffusers to generate uniform lighting. Two Ulanzi L2 Lite (5500 K) co-moving lights were mounted on a custom-built, externally tracked underwater camera (Winkel et al. 2023) to emulate active underwater lighting. Representative images taken from the tank are shown in Fig. 3, and Table 1 provides an overview of the datasets. Small-scale test scenes were captured using the camera, with external tracking serving as ground truth. To ensure accurate pose estimation underwater, the camera was mounted on a stick equipped with two HTC Vive controllers, allowing precise measurement of position and orientation in air while taking images underwater, as shown in Figure 2. This data was fused to estimate the underwater camera trajectory, achieving mean tracking errors below 3 mm and $0.3°$ for translation and rotation, respectively. The camera used a fisheye lens with a minimum working distance of 300 mm and an angle of view of $180°$ diagonal, $143°$

**Table 2.** COLMAP reconstruction statistics for datasets E1–E14, including registered images, 3D points, observations, mean track length, observations per image, reprojection error, and number of submaps.

| Dataset | Registered Images | Points | Observations | Mean Track Length (Avg. 2D Obs. per 3D points) | Mean Obs. per Img. | Mean Reproj. Error (px) | Num. Submaps |
|---|---|---|---|---|---|---|---|
| E1 | 1688 (/1688) | 550,889 | 12,033,387 | 21.84 | 7128.78 | 0.74 | 1 |
| E2 | 1772 (/1772) | 532,113 | 11,808,548 | 22.19 | 6663.97 | 0.79 | 1 |
| E3 | 2404 (/2404) | 571,885 | 13,712,059 | 23.98 | 5703.85 | 0.87 | 1 |
| E4 | 1911 (/1911) | 567,374 | 15,686,130 | 27.65 | 8208.34 | 0.70 | 1 |
| E5 | 1818 (/1818) | 510,067 | 13,236,556 | 25.97 | 7286.33 | 0.76 | 1 |
| E6 | 1704 (/1704) | 542,158 | 12,289,035 | 22.67 | 7211.88 | 0.70 | 1 |
| E7 | 1414 (/1414) | 496,408 | 7,170,896 | 14.44 | 5071.36 | 0.77 | 1 |
| E8 | 1636 (/1636) | 44,662 | 832,163 | 18.63 | 508.657 | 0.84 | 1 |
| E9 | 1497 (/1508) | 59,749 | 550,081 | 9.21 | 367.456 | 1.11 | 1 |
| E10 | 1413 (/1413) | 39,343 | 433,677 | 11.02 | 306.919 | 0.74 | 1 |
| E11 | 1417 (/1591) | 27,891 | 293,390 | 10.52 | 207.05 | 0.79 | 1 |
| E12 | 1716 (/1805) | 38,243 | 451,398 | 11.80 | 263.052 | 0.70 | 3 |
| E12_crop | 1711 (/1805) | 29,828 | 363,892 | 12.20 | 212.678 | 0.69 | 2 |
| E12_crop_clahe | 1796 (/1805) | 127,646 | 1,255,026 | 9.83 | 698.79 | 0.80 | 1 |
| E13 | 668 (/1771) | 11,087 | 134,863 | 12.16 | 201.891 | 0.71 | 6 |
| E14 | 1945 (/1946) | 183,550 | 233,2148 | 12.70 | 1188.05 | 0.79 | 1 |

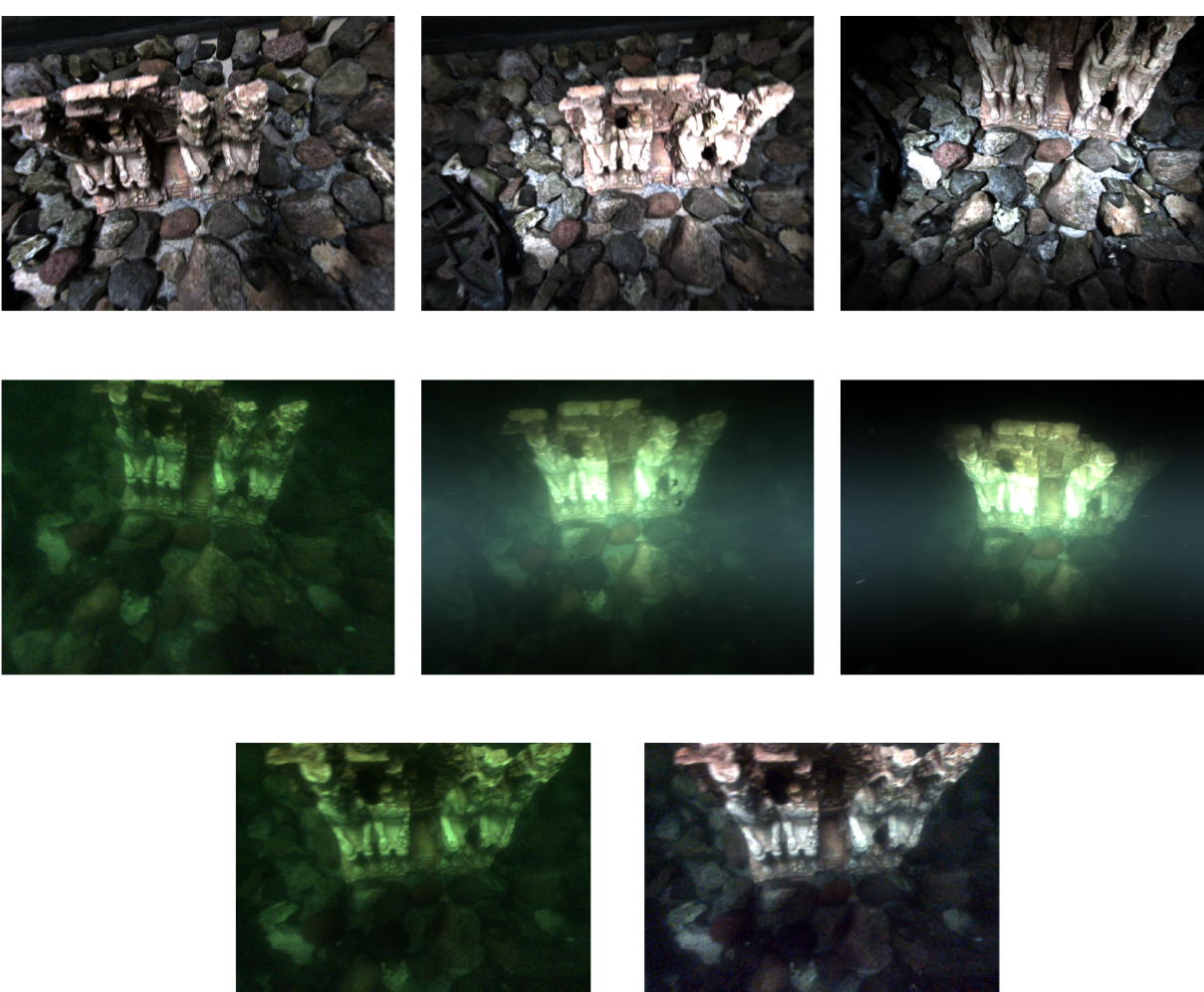

**Figure 3.** Left to right: Example images of tank sets: with Sunlight, Sun-and artificial light and artificial light. Upper row: in air **E**1-3, **E**4-5, and **E**6-7. Middle row: underwater **E**8-9, **E**10-11, and **E**12-13. Lower row: linear sRGB image (left); image after Seathru-NeRF preprocessing (right). Please note that the images are shown in sRGB-space for better visibility, no enhancement is applied

horizontal, and 106° vertical. We then employ the following calibration scheme, to determine the extrinsics and intrinsics of the camera for the in-air sets (E1-E7). Initially, fisheye calibration was performed in air with a residual error of 0.22 px. Next, we followed the approach in(She et al. 2019, 2022) to obtain and correct the camera's displacement with respect to the dome center. This step ensures that refraction effects due to the traversal of the light rays through interfaces between media with different optical densities are omitted. For underwater datasets (E8–E14), a separate fisheye calibration is carried out in a clear water setting, reaching a residual error of 0.55 px. The last step and the dome centering are carried out to eliminate all refraction-based effects such that the underwater sets can be used for pure radiometric problems. The image sequences captured in air include uniformly lit sequences (E1–E3), mixed lighting (E4–E5), and co-moving lights (E6–E7), following lawn mower or free 3D scanning trajectories. Water was then added along with dye to emulate the greenish color typical of the Baltic Sea, and Maalox (Laux et al. 2002) was used as a scattering agent to reproduce realistic underwater visibility effects The underwater sequences mimic sunlight conditions (E8–E9), mixed sun and artificial illumination (E10–E11), and deep-sea scenarios with only artificial lighting (E12–E13). In each pair, the first sequence follows a lawn mower trajectory, while the second uses free trajectories with larger depth variations. Despite the small tank size, this dataset systematically combines controlled lighting, diverse camera trajectories, and varying water conditions to study reconstruction challenges underwater. Unlike most existing underwater datasets, which focus primarily on sunlight conditions, our dataset includes sequences with artificial lighting, mixed illumination, and deep-sea scenarios, enabling evaluation under a wider range of conditions. While scaling this dataset to larger tanks or real-world environments would require careful control of lighting, tracking, and water properties, the proposed methodology provides a practical blueprint for constructing comparable datasets at larger scales. The controlled setup enables repeatable benchmarking while yielding insights relevant to real-world underwater scenarios. As summarized in Table 1, each sequence begins with a short initialization maneuver consisting of a small "wiggle" over a single point, followed by a lawn-mower scanning pattern and a final track-crossing segment to improve loop closure. Free trajectories correspond to free-flying scans, which introduce more loop-closing opportunities but also larger height variations that can degrade loop closure in scattering media due to strong appearance changes.

## *Code and Data Availability*

The BALTIC dataset, evaluation toolkit, and reconstruction utilities are publicly available to ensure full reproducibility.

**Dataset.** The full dataset (E1–E13) is hosted on Hugging Face: `https://huggingface.co/datasets/Michele1996/BALTIC`

It can be downloaded programmatically as:

```
snapshot_download("Michele1996/BALTIC",
                  repo_type="dataset",
                  local_dir="BALTIC")
```

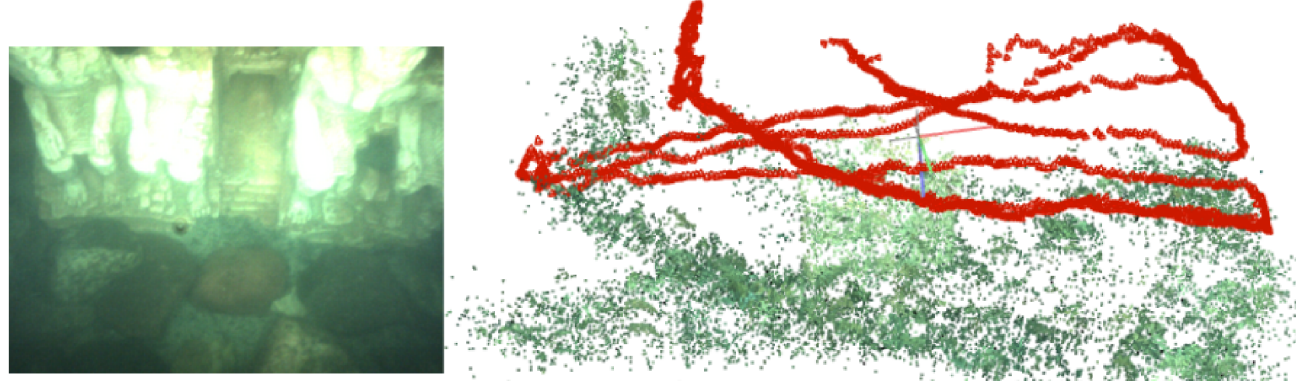
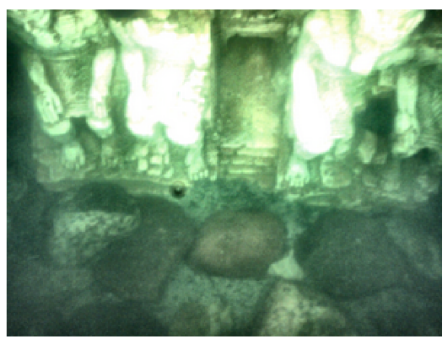
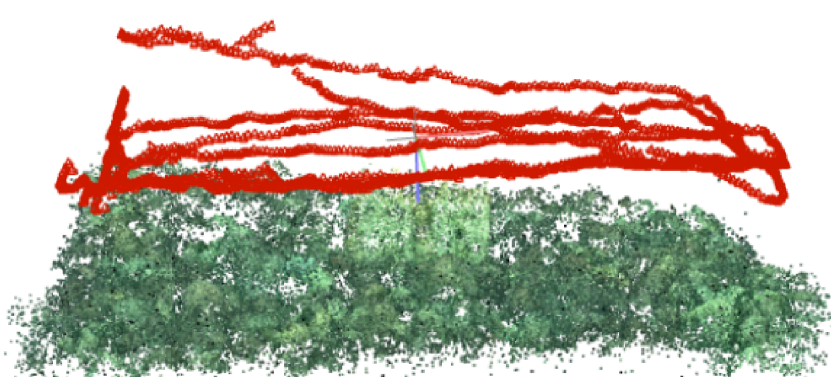

**Figure 4.** E12 dataset with preprocessing. Left: image cropping to focus on the central region and suppress light cones. Right: cropping with CLAHE enhancement, yielding denser and more accurate trajectory reconstruction.

**Code and devkit.** All reconstruction and evaluation scripts are available at: `https://github.com/Michele1996/baltic-project-page`

The repository provides a lightweight devkit for reproducible evaluation, including:

- Geometry evaluation from COLMAP reconstructions (Chamfer distance, surface roughness, nearest-neighbor spacing)
- Trajectory evaluation under similarity alignment (Absolute Trajectory Error, Relative Pose Error)
- Parsers for reconstructed trajectories (`E*_xyz_q.txt`)
- Parsers for ground-truth tracking data (`E*_tf.txt`)

**Reconstruction protocol.** Each sequence is reconstructed using COLMAP. The resulting sparse point clouds and trajectories are then evaluated using the provided devkit.

## Structure-from-Motion Evaluation on the BALTIC Dataset

We evaluate Structure-from-Motion (SfM) using COLMAP on the proposed BALTIC dataset, which includes both in-air and underwater sequences under varying lighting and motion conditions. Our analysis focuses on reconstruction quality, trajectory accuracy, and the impact of cross-domain augmentation. We assess how performance degrades across conditions and identify factors that influence robustness in underwater scenarios.

### *Sparse reconstruction and registration performance*

We evaluate COLMAP on all datasets (E1–E13) using linear sRGB images to analyze camera registration and sparse scene reconstruction performance. COLMAP's exhaustive matcher reliably selects initial image pairs for in-air datasets (E1–E7), while underwater datasets (E8–E13) often require the sequential matcher due to fewer feature correspondences. While SfM performs reliably in in-air sequences, its performance degrades significantly underwater due to reduced feature visibility and photometric inconsistencies, as analyzed in the following. Representative reconstructions are shown in Fig. 5, and Table 2 summarizes reconstruction statistics, including registered images, recovered 3D points, track lengths, and reprojection errors. In air sequences (E1–E7), COLMAP successfully registered most frames, producing dense reconstructions with long feature tracks, particularly for structured trajectories (E1, E2, E4, E6). Free-flight trajectories (E3, E5, E7) yielded fewer points and slightly higher reprojection errors, while artificial lighting (E4–E7) improved track lengths and point density. Underwater sequences (E8–E13) show a significant drop in reconstruction quality. Registered points decrease by an order of magnitude, tracks are shorter, and several reconstructions fragment into multiple submaps (e.g., E12, E13), reflecting fewer reliable correspondences and reduced global consistency. Initialization from well-matched image pairs (E2, E4, E6, E8, E10, E12) improves registration and reduces reprojection errors, even underwater. We further studied the effect of image preprocessing on dataset E12, which originally produced three submaps. Cropping to the main light cone improved trajectory accuracy and reduced fragmentation to two submaps. Applying CLAHE (Pizer et al. 1987) to the cropped images increased point cloud density and trajectory fidelity, producing a single cohesive map (see Fig. 4), in line with (Summers et al. 2025). This shows that focusing on well-lit regions and enhancing contrast can partially mitigate underwater degradation. While COLMAP remains robust in air under controlled conditions, underwater reconstruction remains challenging, highlighting the need for stronger priors or learned models to address scattering, absorption, and color degradation.

**Table 3.** Mesh statistics for 14 datasets. Surface areas and average curvatures are rounded to two decimals. Metrics highlighted in orange indicate values closest to those of E1.

| Dataset | Triangles | Surface Area (cm$^2$) | Avg Curvature (1/cm) |
|---|---|---|---|
| E1 | 1937032 | 20187.76 | 0.07 |
| E2 | **1937363** | 19317.44 | **0.07** |
| E3 | 1659467 | 33823.47 | 0.10 |
| E4 | 1536374 | 20319.91 | 0.08 |
| E5 | 1111556 | 30595.40 | 0.12 |
| E6 | 1356240 | **20226.32** | 0.09 |
| E7 | 1154816 | 37542.58 | 0.12 |
| E8 | 728756 | 201086.62 | 0.18 |
| E9 | 692594 | 149024.19 | 0.17 |
| E10 | 489263 | 31281.49 | 0.16 |
| E11 | 431599 | 45581.46 | 0.20 |
| E12 | 428189 | 32163.55 | 0.17 |
| E13 | 311656 | 31443.70 | 0.18 |
| E14 | 314748 | 21520.18 | 0.17 |

### *Trajectory Accuracy Evaluation*

We quantitatively evaluate COLMAP reconstructions by comparing estimated camera trajectories to ground truth poses from the HTC Vive system. Trajectories were first aligned using the Umeyama method (Umeyama 1991). Figure 6 reports Absolute Trajectory Error (ATE) and Mean Relative Pose Error (RPE) for all 14 datasets (E1–E13). Air sequences (E1–E7) generally achieve low RPE, indicating precise local pose estimates, while ATE varies with trajectory length and scene complexity. Underwater sequences (E8–E13) exhibit higher ATE and RPE, reflecting larger global and local deviations due to scattering, absorption, and reduced feature visibility. Datasets (E1–E14) show very low RPE (0.002–0.016m) but moderate ATE (0.08–0.4m), indicating that local motions are estimated with high precision, while small errors accumulate over time to produce some global trajectory drift. We also include E14, a cross-domain reconstruction where the underwater sequence E12 is augmented with a minimal set of in-air views from E6. Since all reconstructions are evaluated against the respective ground truth, this setting allows us to directly assess whether adding cross-domain information improves not just visual and geometric detail (Figure 1) but also trajectory stability. Indeed, E14 demonstrates that complementary in-air views can reduce drift and improve alignment, reinforcing our central claim that cross-domain information is both practical and beneficial. These results confirm that COLMAP performs well in well-lit air conditions but struggles underwater, while cross-domain reconstructions offer a promising avenue to mitigate these challenges, complementing qualitative observations from point cloud density and reconstruction consistency. Table 4 reports point cloud alignment metrics: Scale RMS measures global size consistency, ICP Fitness quantifies rigid alignment quality, Chamfer RMS is the root mean square distance between corresponding points in two clouds (capturing geometric deviation), Surface Roughness reflects local point dispersion, and Mean NN Distance indicates point density. Air datasets maintain Scale RMS near 1, high ICP fitness, and low Chamfer RMS and roughness, confirming accurate and smooth reconstructions. Underwater sequences exhibit lower ICP fitness and higher Chamfer RMS and roughness, consistent with larger deviations. The cross-domain reconstruction E14 improves Chamfer RMS and nearest-neighbor distances relative to E12, showing that complementary in-air views reduce drift and enhance geometric fidelity.

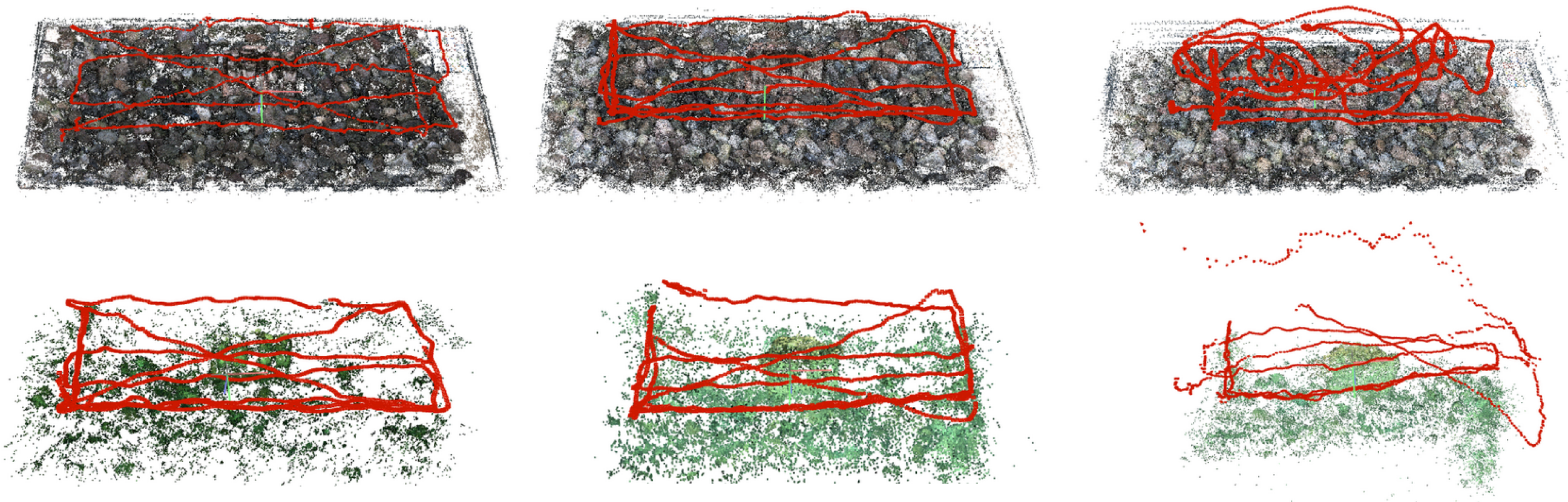

**Figure 5.** Representative COLMAP reconstructions for six datasets: E1, E4, E6 (air) and E8, E10, E12 (underwater). E1 and E8 were captured under sunlight only, E4 and E10 combine sunlight with artificial illumination, and E6 and E12 use purely artificial lighting in air and underwater, respectively. Air reconstructions remain dense and globally consistent, whereas underwater results appear sparser and fragmented due to scattering and absorption effects.

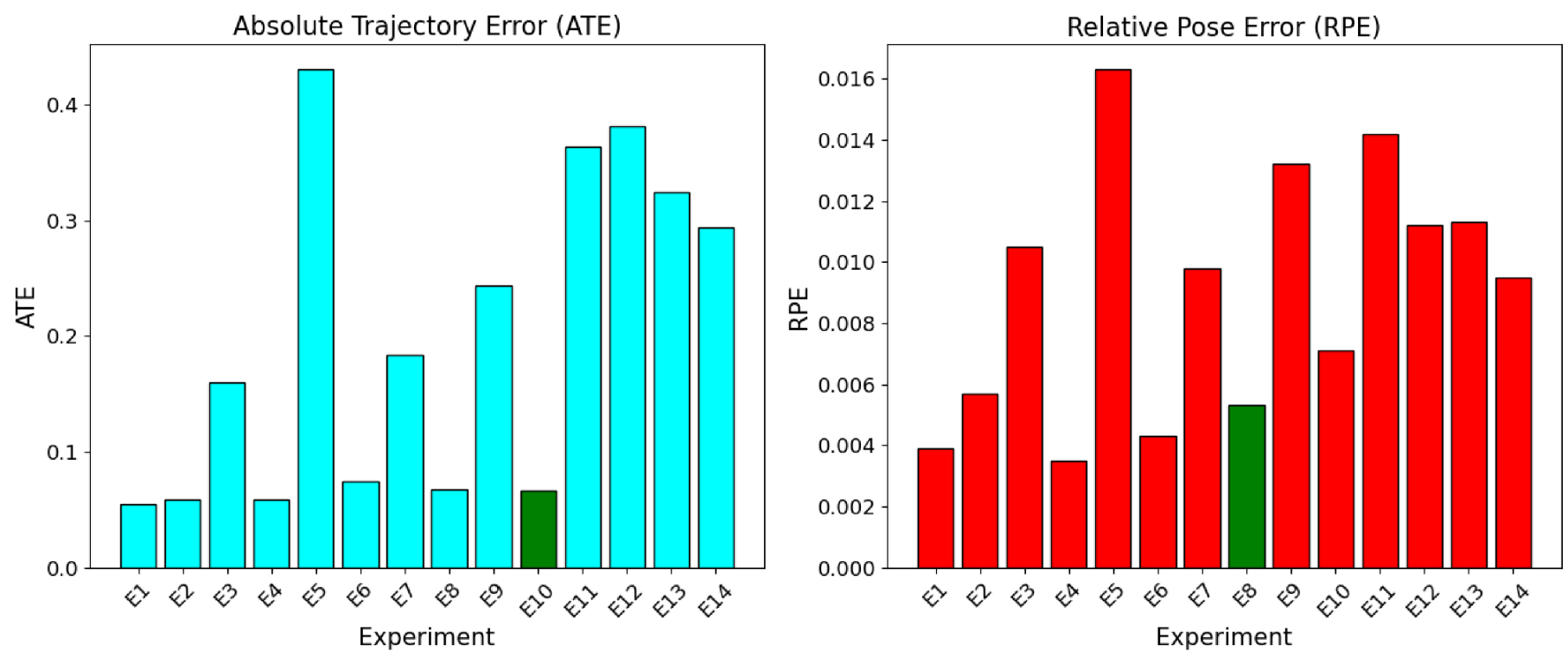

**Figure 6.** Absolute Trajectory Error (ATE) and Relative Pose Error (RPE) expressed in centimeters across all experiments. The lowest error for the underwater sets (E8–E13) is highlighted in green. Cross-domain reconstruction (E14) shows a slight improvement in ATE and RPE compared to the pure underwater set E12. This suggests that while the mixed reconstruction can appear visually correct, the air viewpoints may not provide sufficient coverage for achieving a highly accurate trajectory.

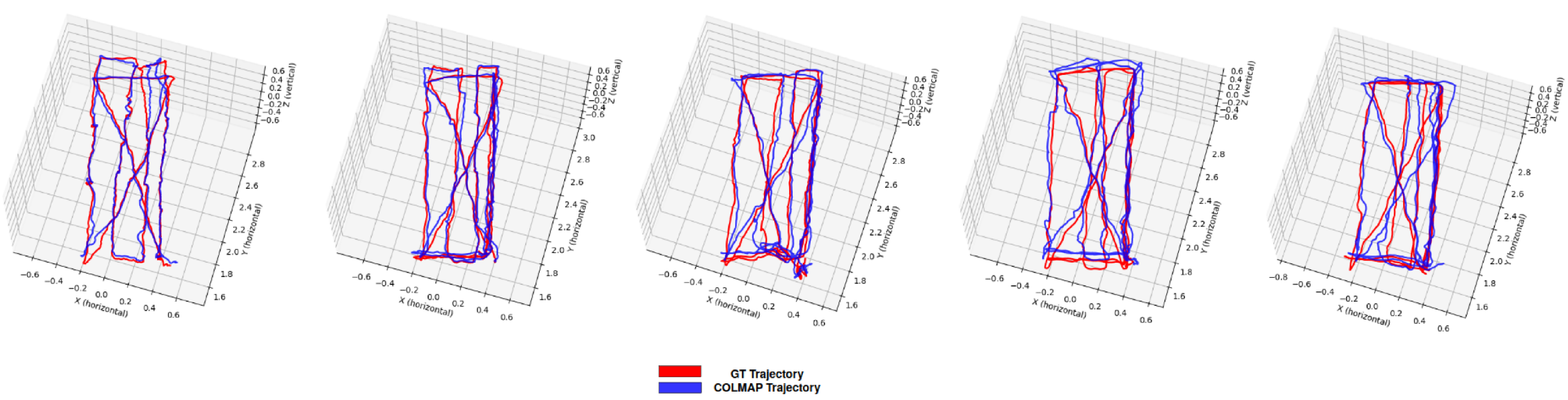

**Figure 7.** Comparison of COLMAP trajectories against ground truth for six representative datasets, both expressed in meters: E1, E4, E6 (air) and E8, E10 (underwater). Air trajectories remain dense and closely follow ground truth, while underwater trajectories show larger deviations due to reduced feature visibility.

**Table 4.** Point cloud alignment metrics for air (E2–E7) and underwater (E8–E14) sequences. Each dataset is scaled using RMS-based normalization to the reference point cloud from E1 and refined with rigid ICP.

| Dataset | Scale_RMS | ICP_Fitness | Chamfer_RMS_mm | Surface_Roughness | Mean_NN_Distance_mm |
|---|---|---|---|---|---|
| E2 | 1.0055 | 0.5197 | 200.30 | 2.388e-06 | 14.82 |
| E3 | 0.8738 | 0.1681 | 497.83 | 2.679e-06 | 18.24 |
| E4 | 0.9717 | 0.4529 | 296.93 | 1.730e-06 | 15.27 |
| E5 | 0.8761 | 0.0758 | 1109.38 | 1.630e-06 | 18.41 |
| E6 | 0.9995 | 0.2479 | 454.72 | 1.847e-06 | 16.52 |
| E7 | 0.8726 | 0.0434 | 593.79 | 1.737e-06 | 21.19 |
| E8 | 0.3172 | 0.0085 | 1847.86 | 4.048e-06 | 33.67 |
| E9 | 0.6447 | 0.2115 | 1386.61 | 1.857e-06 | 47.06 |
| E10 | 1.0800 | 0.0423 | 870.30 | 5.254e-07 | 41.49 |
| E11 | 0.9277 | 0.0041 | 1185.85 | 1.539e-06 | 48.94 |
| E12 | 1.1608 | 0.0117 | 892.41 | 7.356e-07 | 43.68 |
| E13 | 0.9114 | 0.0117 | 951.73 | 4.717e-07 | 47.66 |
| E14 | 1.0943 | 0.0840 | 783.34 | 1.444e-06 | 33.32 |

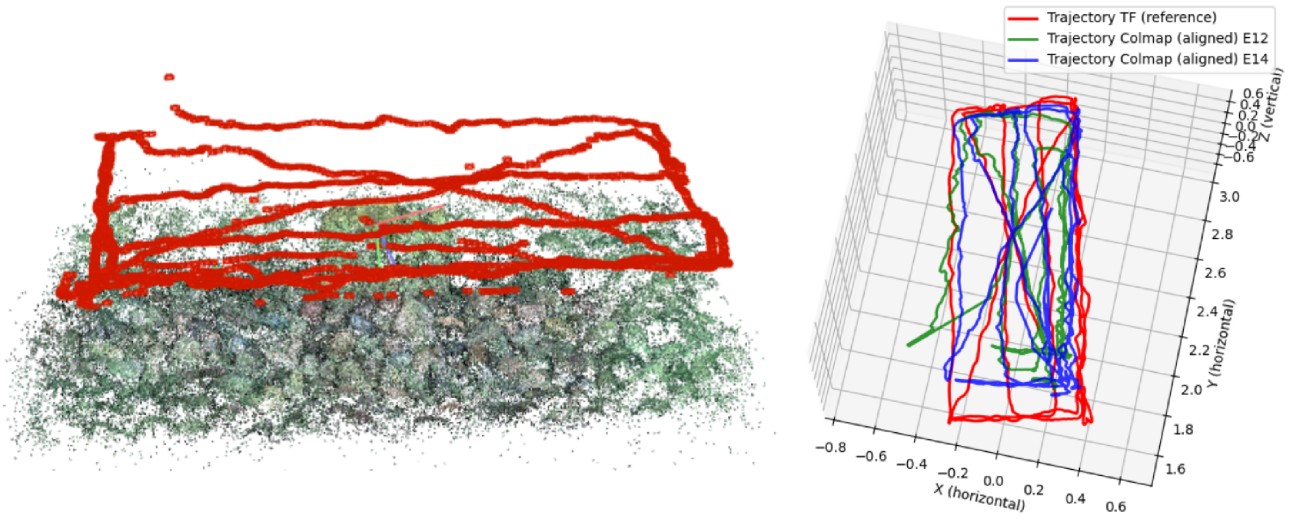

**Figure 8.** Cross-domain reconstruction for dataset E12 by augmenting the underwater sequence with an optimal subset of in-air images captured under the same artificial lighting conditions. Trajectories are expressed in meters.

## *Dense Reconstruction*

We evaluated the dense 3D reconstruction of the scenes using COLMAP's Delaunay mesh outputs. For each dataset (E1–E14), we compute key geometric metrics from the reconstructed meshes, including the number of triangles, surface area, and average curvature. The meshes are scaled to a common metric reference to allow a meaningful comparison. Table 3 summarizes the statistics computed. E1 serves as the reference reconstruction, representing the baseline for comparison. The closest datasets to E1 in each metric is highlighted in yellow. The results show that the number of triangles varies between datasets, reflecting differences in reconstruction completeness and scene coverage. The surface area generally correlates with the amount of detail captured in the scene. Compared to E12, E14 has a slightly lower surface area but similar average curvature, indicating that the cross-domain reconstruction preserves overall geometric quality while being slightly more compact. Average curvature provides insight into local geometric smoothness, with lower values observed in in-air reconstructions and higher values in underwater or cross-domain datasets, suggesting more variation or potential reconstruction artifacts.

## *Cross-Domain Reconstruction*

We further explored radiometric cross-domain reconstruction by augmenting underwater datasets with a minimal subset of in-air images captured under similar lighting conditions. In this setting, geometry is assumed to be already resolved, as all images are undistorted into a common geometric space that proved consistent in our case. The goal was instead to identify the smallest set of in-air views that provide maximal radiometric support to weak regions of the underwater reconstruction. Formally, we partition the underwater point cloud $\mathcal{P}_{uw}$ into voxels of size $v = 0.02$ m and define the set of weak voxels $\mathcal{V}_w = \{v_i \mid |P(v_i)| < 3\}$, where $P(v_i)$ denotes the number of reconstructed points in voxel $v_i$. This voxel size and threshold were chosen to balance spatial resolution with robustness to sparsely reconstructed regions, ensuring weak areas are effectively identified without over-fragmenting the point cloud. Each in-air image $I_j$ is scored by the number of weak voxels it observes:

$$s(I_j) = \sum_{v_i \in \mathcal{V}_w} \mathbf{1}\big[\pi_j(c(v_i)) \in \Omega_j\big],$$

where $c(v_i)$ is the center of the voxel, $\pi_j$ the projection under the camera model of $I_j$, and $\Omega_j$ the image domain. A greedy selection is applied, iteratively choosing the image that maximizes new weak-voxel coverage. This approach efficiently prioritizes images that contribute the most additional coverage, avoiding unnecessary redundancy while keeping computation manageable. Applied to sequence E12, which originally produced multiple fragmented submaps, this strategy selected **144 in-air images**, providing near-complete coverage and enabling COLMAP to fuse underwater and in-air data into a single dense and globally consistent map (Fig. 8), in contrast to the fragmented three-submap reconstruction observed with underwater images alone. Sequence E12 was chosen for illustration because its fragmented reconstruction provides a clear scenario to demonstrate the benefits of targeted cross-domain augmentation. These results indicate that carefully targeted radiometric augmentation across domains can compensate for scattering and absorption, and represents a promising avenue for improving underwater reconstructions once geometry is normalized. In real underwater settings, object appearance may differ from in-air captures due to scattering, absorption, or temporal effects such as corrosion or biofouling. While this method is highly effective for man-made objects (e.g., artifacts or pipes), its applicability to natural rocks or objects covered in sea grass, rust, or biofouling can be limited by occlusion, surface changes, or low texture. Nonetheless, our benchmark provides guidance on minimal image selection and coverage, and robust features extracted from in-air images can be reprojected into underwater views to provide geometric anchors when appearance degrades, complementing radiometric augmentation and extending cross-domain reconstruction to more challenging real-world conditions.

| Dataset | Nerfacto | | | SeaThru-NeRF | | | Gaussian Splatting | | | WaterSplat | | | SeaSplat | | |
|---|---|---|---|---|---|---|---|---|---|---|---|---|---|---|---|
| | PSNR ↑ | SSIM ↑ | LPIPS ↓ | PSNR ↑ | SSIM ↑ | LPIPS ↓ | PSNR ↑ | SSIM ↑ | LPIPS ↓ | PSNR ↑ | SSIM ↑ | LPIPS ↓ | PSNR ↑ | SSIM ↑ | LPIPS ↓ |
| E1 | 21.12 | 0.66 | 0.23 | 24.31 | 0.78 | 0.23 | **30.81** | **0.95** | **0.06** | 26.68 | 0.85 | 0.18 | 26.95 | 0.89 | 0.12 |
| E2 | 21.03 | 0.67 | 0.22 | 23.83 | 0.78 | 0.24 | **30.34** | **0.94** | **0.07** | 26.48 | 0.86 | 0.18 | 26.94 | 0.90 | 0.12 |
| E3 | 21.86 | 0.70 | 0.23 | 25.09 | 0.81 | 0.23 | **28.85** | **0.93** | **0.10** | 26.00 | 0.85 | 0.20 | 26.54 | 0.89 | 0.14 |
| E4 | 23.34 | 0.78 | 0.19 | 20.95 | 0.78 | 0.23 | **28.19** | **0.95** | **0.06** | 24.93 | 0.88 | 0.14 | 23.44 | 0.88 | 0.12 |
| E5 | 19.68 | 0.74 | 0.21 | 18.85 | 0.72 | 0.28 | **25.00** | **0.91** | **0.09** | 21.59 | 0.83 | 0.19 | 23.14 | 0.86 | 0.14 |
| E6 | 21.51 | 0.72 | 0.22 | 17.86 | 0.67 | 0.32 | **26.76** | **0.91** | **0.10** | 23.45 | 0.84 | 0.19 | 20.06 | 0.70 | 0.29 |
| E7 | 16.22 | 0.57 | 0.32 | 14.50 | 0.53 | 0.48 | **22.34** | **0.83** | **0.18** | 19.56 | 0.75 | 0.30 | 17.57 | 0.63 | 0.34 |
| E8 | 19.10 | 0.26 | 0.97 | 18.83 | 0.26 | 0.92 | **23.36** | **0.48** | **0.42** | 20.89 | 0.28 | 1.00 | 20.84 | 0.36 | 0.50 |
| E9 | 18.36 | 0.26 | 0.96 | 18.65 | 0.27 | 0.91 | **23.67** | **0.49** | **0.43** | 21.15 | 0.29 | 0.99 | 21.11 | 0.37 | 0.51 |
| E10 | 25.70 | 0.87 | 0.26 | 17.35 | 0.76 | 0.49 | **32.09** | **0.93** | **0.26** | 30.77 | 0.91 | 0.29 | 20.77 | 0.75 | 0.48 |
| E11 | 16.94 | 0.63 | 0.51 | 13.73 | 0.56 | 0.61 | **22.48** | **0.75** | **0.41** | 21.35 | 0.74 | 0.53 | 15.75 | 0.58 | 0.57 |
| E12* | 20.11 | 0.69 | 0.45 | 16.15 | 0.62 | 0.54 | 22.70 | 0.76 | **0.40** | **24.14** | **0.78** | 0.48 | 20.41 | 0.60 | 0.53 |
| E13* | 21.14 | 0.68 | 0.44 | 17.20 | 0.62 | 0.54 | 24.68 | 0.78 | **0.37** | **25.68** | **0.78** | 0.46 | 15.77 | 0.56 | 0.57 |
| E14 | 28.03 | 0.82 | 0.41 | 23.16 | 0.78 | 0.47 | 27.26 | **0.87** | **0.40** | **29.98** | 0.85 | 0.42 | 18.18 | 0.66 | 0.62 |

**Table 5.** Quantitative results on 13 datasets using PSNR, SSIM, and LPIPS metrics across five methods. For each dataset, the best-performing method per metric is highlighted. Gaussian Splatting achieves the best results in most cases, except for E12 and E13 where WaterSplat performs better. The * denotes evaluation on the most complete submap.

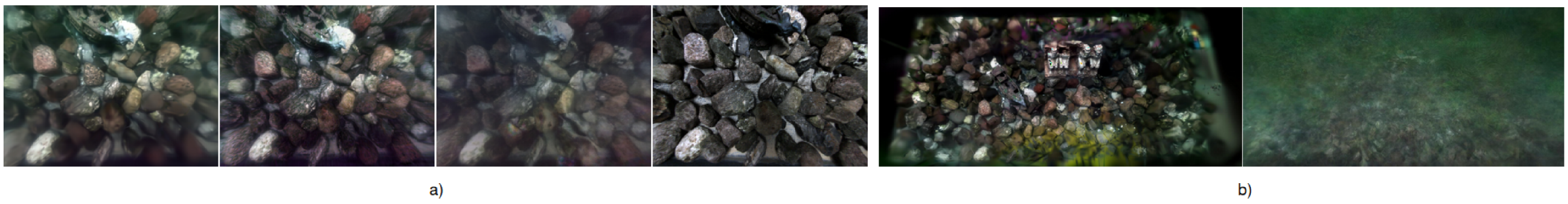

**Figure 9.** (a) Water-removed renderings produced by WaterSplat (left), then SeaSplat, SeaThru-NeRF and ground-truth (right). (b) Water-removed surface meshes reconstructed with WaterSplat (left) and SeaThru-NeRF (right). SeaSplat does not provide a 3D mesh representation for the water-removed reconstruction. All the images/mesh are from E8 dataset.

## Rendering Performance

We evaluated reconstruction quality across all 13 datasets using five state-of-the-art methods: Nerfacto, SeaThru-NeRF, Gaussian Splatting (basic 3DGS), WaterSplat, and SeaSplat. All methods used identical preprocessing following SeaThru-NeRF, consisting of exposure normalization and white balance correction to ensure consistent input appearance. Exposure normalization was estimated from image timestamps and used to rescale intensities as if captured under uniform exposure. Figure 3 shows an example for dataset E8, while Table 5 reports PSNR, SSIM, and LPIPS scores. Air datasets (E1–E7) generally exhibit higher and more consistent performance due to dense and globally coherent COLMAP reconstructions. An exception is E7, which shows degraded performance across all methods caused by the combination of limited illumination and increased camera motion, rather than the trajectory type itself. Gaussian Splatting consistently achieves the highest PSNR and SSIM across these datasets, with notably low LPIPS values, indicating accurate structure and appearance reproduction. Nerfacto and SeaThru-NeRF perform reasonably well under sunlight-only conditions (E1, E3) and mixed illumination (E4), though their LPIPS scores suggest slightly lower perceptual fidelity compared to splatting-based methods. WaterSplat and SeaSplat produce competitive results, with WaterSplat showing particular robustness under artificial lighting (E6). Underwater datasets (E8–E13) show markedly lower PSNR and SSIM overall, reflecting the sparse and fragmented COLMAP reconstructions caused by scattering and absorption. Nerfacto and SeaThru-NeRF struggle under these conditions, particularly on E8 and E9, with LPIPS values approaching 1, indicating perceptual degradation. Gaussian Splatting maintains moderate reconstruction quality but can exhibit noticeable artifacts where input geometry is weak. WaterSplat and SeaSplat provide improved robustness in scenes with partial or structured lighting (E10, E12), though their performance remains below that observed in air datasets. Illumination effects are clear. In air, sunlight-only datasets (E1, E3) achieve slightly higher

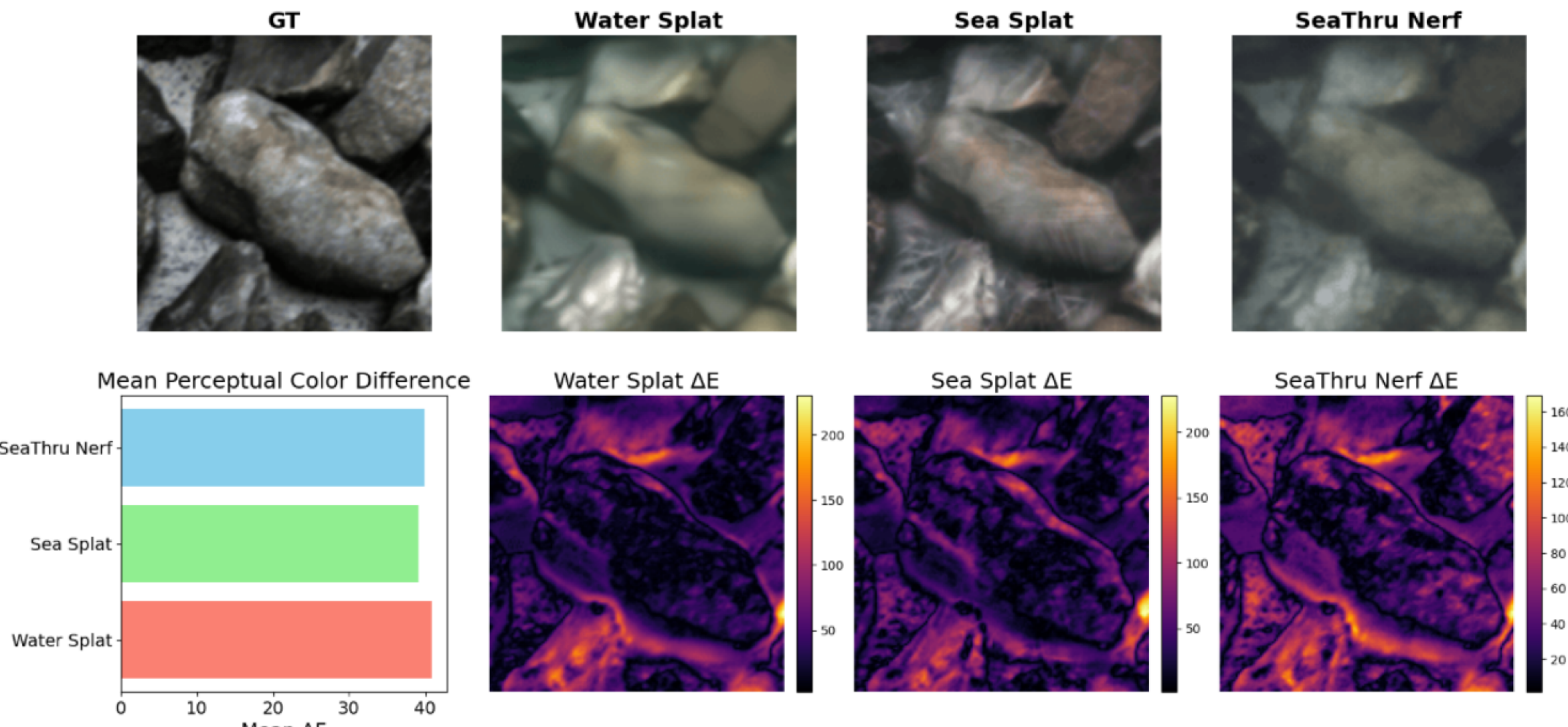

**Figure 10.** Top row: cropped regions from the original images (GT and methods). Bottom row: mean perceptual color difference (left) and $\Delta E$ heatmaps computed in CIE Lab space for each method.

**Table 6.** Representative color difference metrics (CIE Lab) for dataset E8. Analysis performed on a partial region to ensure comparable viewpoints.

| Method | $\Delta E$ Mean | $\Delta E$ Std | $\Delta E$ Min | $\Delta E$ Max | $L$ Diff | $a$ Diff | $b$ Diff |
|---|---|---|---|---|---|---|---|
| WaterSplat | 40.86 | 31.53 | 2.45 | 230.04 | 21.63 | -4.11 | 3.94 |
| SeaSplat | 39.11 | 28.97 | 1.41 | 228.01 | 9.28 | 0.58 | -0.15 |
| SeaThru-NeRF | 39.88 | 25.28 | 1.41 | 169.03 | 6.69 | -2.02 | -0.69 |

PSNR and SSIM compared to purely artificial lighting (E6), suggesting that natural illumination aids feature matching and geometry recovery. Mixed illumination (E4, E10) can improve overall visibility but introduces subtle inconsistencies, reflected in slightly higher LPIPS scores. Underwater, lighting effects are amplified: E10 and E12, which include artificial lights, consistently outperform sunlight-only underwater datasets (E8, E9), demonstrating that controlled lighting helps mitigate scattering and color attenuation. For E8, we additionally evaluated the effect of white balance preprocessing. Applying white balance, following SeaThru-NeRF (Levy et al. 2023), increased PSNR from 15.05 to 18.83 but reduced SSIM from 0.35 to 0.25 and LPIPS from 0.76 to 0.92, indicating improved pixel-wise fidelity but degraded perceptual and structural consistency. Figure 10 provides a qualitative comparison on dataset E8, showing water-removed renderings and reconstructed meshes for WaterSplat, SeaSplat, and SeaThru-NeRF, highlighting differences in appearance restoration and geometric completeness under scattering conditions. Preprocessing also led to faster convergence, reducing training time by about 15 minutes (from 1h15 to 1h). Although throughput and frame rate remained largely unchanged (200k rays/sec, 0.41 FPS), we do not report these metrics in the main table, as our focus is on reconstruction quality. White balance preprocessing on E8 further illustrates that even simple color correction can significantly improve reconstruction metrics and reduce training time. Relative to E12 alone, E14 achieves higher PSNR and SSIM while lowering LPIPS, showing that even limited cross-domain input can stabilize geometry and improve perceptual fidelity. Our results highlight the benefit of fusing air and underwater views to mitigate degradation in purely underwater reconstructions.

We also show that Gaussian Splatting performs consistently well across air, underwater, and cross-domain settings with only basic preprocessing, suggesting that standard methods can remain effective under diverse acquisition conditions. However, commonly used metrics such as PSNR, SSIM, and LPIPS emphasize visual fidelity and may not fully capture geometric completeness or robustness in challenging scenarios.

### *Color Restoration Evaluation*

To further evaluate perceptual color accuracy, we computed CIE Lab differences between the reconstructed images and ground truth for a representative subset of E8. The analysis was performed on a partial region where corresponding viewpoints were sufficiently aligned to ensure meaningful comparison. Table 6 reports per-method $\Delta E$ statistics, including mean, standard deviation, minimum, maximum, and channel-wise ($L$, $a$, $b$) differences. WaterSplat exhibits the highest mean $\Delta E$ (40.86) with substantial brightness adjustments ($L$ mean 21.63), while SeaSplat shows slightly lower $\Delta E$ (39.11) with more balanced corrections. SeaThru-NeRF achieves intermediate $\Delta E$ (39.88) with lower peak deviations (Max 169.03) and subtler chromatic shifts. WaterSplat and SeaSplat perform stronger perceptual color corrections, restoring contrast and chromatic information under underwater scattering conditions, whereas SeaThru-NeRF provides smoother color transitions and maintains moderate perceptual fidelity, consistent with the qualitative renderings in Figure 10.

## Conclusion

In this work, we introduced BALTIC, an open-source benchmark for the systematic evaluation of 3D reconstruction methods across air and underwater domains under controlled variations in lighting, motion, and medium. By combining accurate ground-truth trajectories with diverse acquisition conditions, the benchmark enables a detailed analysis of the robustness of modern reconstruction pipelines in scenarios relevant to marine robotics and underwater inspection. Our experimental evaluation highlights several key findings. First, classical Structure-from-Motion pipelines such as COLMAP remain highly reliable in in-air conditions but degrade significantly underwater, where reduced contrast, scattering, and photometric inconsistencies lead to sparse and fragmented reconstructions. Second, while learning-based approaches such as NeRF and 3D Gaussian Splatting benefit from improved representations, their performance remains strongly dependent on the quality of SfM initialization and input consistency. Third, simple preprocessing strategies, including contrast enhancement and white balance correction, can partially mitigate underwater degradation, but do not fully address the underlying loss of visual information. A central insight of this work is the effectiveness of cross-domain reconstruction. We show that augmenting underwater sequences with a small number of in-air images captured under similar lighting conditions provides both geometric and radiometric priors, improving reconstruction completeness and trajectory stability. This result suggests that, in practical deployments, pre-acquisition of in-air data can serve as a lightweight and effective strategy to enhance underwater reconstruction without requiring complex physical models or retraining. Despite these advances, several limitations remain. The proposed benchmark is based on a controlled tank environment, which, while enabling repeatability, does not fully capture the scale, dynamics, and variability of real-world underwater scenes. Additionally, our evaluation focuses primarily on geometric and photometric consistency, without incorporating full radiometric calibration or physically-based image formation models. The effectiveness of cross-domain reconstruction may also depend on object type and scene stability, particularly in the presence of biofouling, occlusions, or temporal changes. Future work will extend this benchmark to larger-scale and in-situ underwater environments, incorporate more diverse water conditions, and explore tighter integration between geometric reconstruction and physics-based rendering models. Furthermore, learning-based approaches that explicitly account for scattering and medium properties represent a promising direction for improving robustness in challenging conditions. BALTIC provides a reproducible framework for systematically analyzing the limitations of current 3D reconstruction methods and demonstrates that cross-domain augmentation is a simple yet effective strategy for improving performance in degraded visual environments.